\pdfoutput=1
\documentclass[11pt]{article}
\usepackage[table]{xcolor}
\usepackage[]{acl}

\usepackage{times}
\usepackage{latexsym}
\usepackage{booktabs}
\usepackage{hyperref}
\usepackage{float}
\usepackage{xcolor}
\usepackage[linesnumbered,ruled,vlined]{algorithm2e}

\SetCommentSty{mycommfont}
\usepackage[T1]{fontenc}
\usepackage[utf8]{inputenc}

\usepackage{microtype}

\usepackage{inconsolata}
\usepackage{multirow}
\usepackage{graphicx}
\usepackage{utfsym}
\usepackage{scalerel}

\newcommand*{\emojiexample}{\scalerel*{\includegraphics{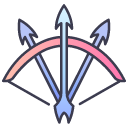}}}%
\title{ Archer \emojiexample): A Human-Labeled Text-to-SQL Dataset with \\Arithmetic, Commonsense and Hypothetical Reasoning  }

\author{Danna Zheng$^{1}$, Mirella Lapata$^{1}$, Jeff Z. Pan$^{1, 2}$ \\
            $^{1}$ School of Informatics, University of Edinburgh, UK\\
            $^{2}$ Huawei  Edinburgh Research Centre, CSI, UK\\
        dzheng@ed.ac.uk, mlap@inf.ed.ac.uk, http://knowledge-representation.org/j.z.pan/
}
\begin{document}
\maketitle
\begin{abstract}
We present Archer, a challenging bilingual text-to-SQL dataset specific to complex reasoning, including arithmetic, commonsense and hypothetical reasoning. It contains 1,042 English questions and 1,042 Chinese questions, along with 521 unique SQL queries, covering 20 English databases across 20 domains. Notably, this dataset demonstrates a significantly higher level of complexity compared to existing publicly available datasets. %The creation of Archer involved meticulous manual annotation, spanning approximately 300 hours of effort. 
Our evaluation shows that Archer challenges the capabilities of current state-of-the-art models, with a high-ranked model on the Spider leaderboard achieving only 6.73\% execution accuracy on Archer test set. Thus, Archer presents a significant challenge for future research in this field.
\end{abstract}

\section{Introduction}
\begin{figure}[ht]
    \centering
    \includegraphics[width=.46\textwidth]{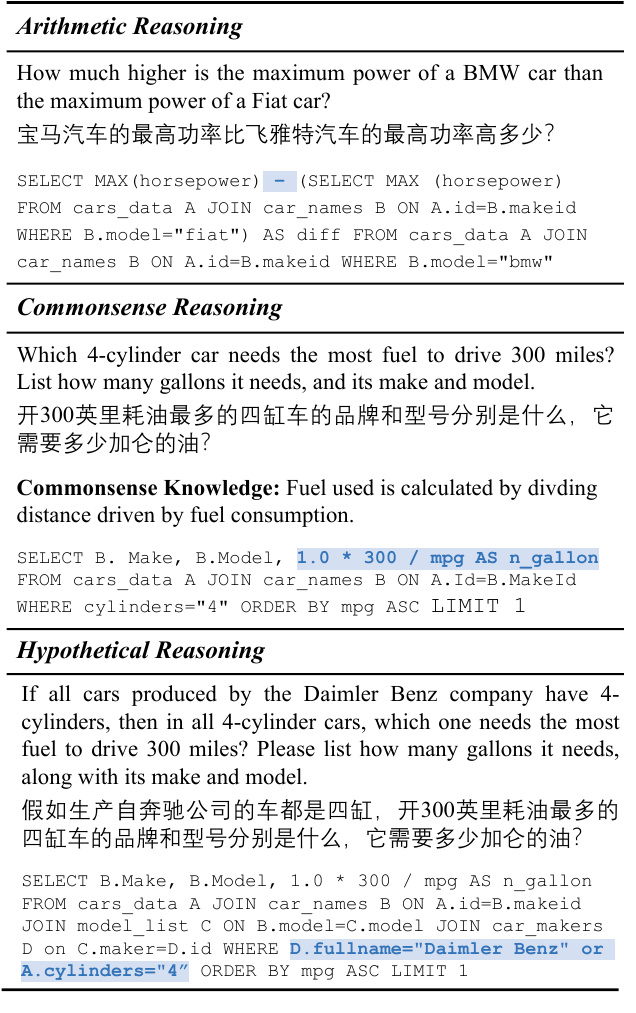}
    \caption{Archer examples with three reasoning types: arithmetic, commonsense, and hypothetical reasoning. (See more examples in Appendix~\ref{app:archer-ex})\vspace{-.3cm}}
    \label{fig:reasoning}
\end{figure}

% Relational databases are widely used for data storage and management due to their robustness and scalability. However, accessing and querying this data often presents difficulties for non-experts and individuals who prefer natural language interfaces, as it necessitates familiarity with structured query languages (SQL). To tackle this challenge, the 
The text-to-SQL task is an important NLP task, which maps  input questions to meaningful and executable SQL queries, 
%focuses on automatically converting natural language queries into SQL queries,
enabling users to interact with databases in a more intuitive and user-friendly manner.
%Significant advancements have been made in the field of text-to-SQL tasks.
%Notably, s
State-of-the-art methods ~\cite{pourreza2023din, li2023resdsql, li2023graphix, scholak-etal-2021-picard} relying on large language models have achieved execution accuracy above 75\% on the Spider dataset\cite{yu-etal-2018-spider}, which encompasses complex SQL grammar and cross-domain settings. Recently, \citet{pourreza2023din} achieved remarkable results with an impressive 85.3\% execution accuracy on the Spider dataset, leveraging the enhanced capabilities of \href{https://openai.com/research/gpt-4}{GPT-4}.
%Considering the substantial progress made by these models, it appears that many of the long-standing issues associated with the text-to-SQL task have been effectively addressed.

However, previous text-to-SQL datasets~\cite{yu-etal-2018-spider,data-sql-advising,data-sql-imdb-yelp,iyer2017learning,zhong2017seq2sql,data-academic,data-restaurants,data-restaurants-original,data-restaurants-logic,dahl-etal-1994-expanding}, %including Spider~\cite{yu-etal-2018-spider}
have limitations that prevent them from capturing complex reasoning effectively.
%Previous datasets mainly aim to cover various SQL grammars and simplify the task of question understanding.
%In these datasets, it's easy to directly derive necessary SQL operations, the schema, and database values from the question itself, without needing additional reasoning or external knowledge. 
For example, Spider~\cite{yu-etal-2018-spider} purposely excludes questions that would require external knowledge, like common-sense inferences or mathematical calculations. This exclusion limits Spider's ability to properly test how well models can handle real-world scenarios, which often require a deeper level of reasoning capabilities.

%To address the need for a more comprehensive and challenging text-to-SQL dataset, 
In this paper, we present Archer,
% (\textbf{Ar}ithmetic, \textbf{C}ommonsense and \textbf{H}ypoth\textbf{e}tical \textbf{R}easoning)
an innovative dataset designed to incorporate three distinct types of reasoning: arithmetic, commonsense, and hypothetical reasoning. By including such varied reasoning skills, Archer seeks to challenge and expand the capabilities of text-to-SQL models, equipping them to manage more intricate and nuanced queries. Figure~\ref{fig:reasoning} showcases data examples from Archer that demonstrate these three reasoning abilities.

To evaluate the challenge posed by Archer, we conducted experiments with both large language models (LLMs) and fine-tuned models. However, all models demonstrated inferior performance when dealing with Archer. Even the model that achieved a high place on the Spider leaderboard managed only 6.73\% execution accuracy on Archer test sets. These findings highlight substantial potential for improvement, indicating that Archer indeed provides a significant challenge to current models.

\section{Reasoning Types}
In this section, we present the three different types of reasoning in Archer: arithmetic, commonsense, and hypothetical reasoning.

\paragraph{Arithmetic reasoning} Arithmetic reasoning pertains to the act of resolving mathematical problems through logical and analytical thought processes.  According to an analysis of SQL queries from practical applications like the Baidu search engine and customer service and data analysis robots by~\citet{wang-etal-2020-dusql}, %revealing that 
mathematical calculations account for a significant portion across SQL applications.  %They found that math calculation occupies a considerable proportion in all applications. 
However, previous high-quality datasets contain very few questions that involve calculations, and %the datasets that do contain a considerable proportion of 
such questions are typically auto-generated with simple grammar. In contrast, all question-SQL pairs included in Archer necessitate arithmetic reasoning and are manually annotated to ensure high quality.

\paragraph{Commonsense reasoning} Commonsense reasoning refers to the capacity to make logical deductions based on implicit knowledge and a broad understanding of how things function in the world. Archer includes questions that necessitate models to comprehend the database, infer missing details, and generate logical inferences to create accurate SQL queries.
As illustrated in Figure~\ref{fig:reasoning}, for the question \textit{"Which 4-cylinder car needs the most fuel to drive 300 miles? List how many gallons it needs, and its make and model."}, the database does not provide an explicit schema about the fuel used to travel 300 miles for each car. It only provides each car's fuel consumption in MPG. Solving this question requires commonsense knowledge, specifically the understanding of \textit{"Fuel used is calculated by dividing distance driven by fuel consumption"} to derive the correct SQL.

\paragraph{Hypothetical reasoning} Hypothetical reasoning takes the complexity a step further, requiring models to have counterfactual thinking ability, which is the ability to imagine and reason over unseen cases based on the seen facts and counterfactual assumptions. Archer includes questions that involve hypothetical situations, requiring the model to understand and reason about conditional relationships. As illustrated in Figure~\ref{fig:reasoning}, consider the hypothetical question \textit{"If all cars produced by the Daimler Benz company have 4-cylinders, then in all 4-cylinder cars, which one needs the most fuel to drive 300 miles? Please list how many gallons it needs, along with its make and model."}.
In this question, the underlying assumption contradicts the factual information stored in the database. The model must comprehend this assumption and convert it into the SQL condition \textit{d.fullname = "Daimler Benz" or a.cylinders = "4"}.

\section{Corpus Construction}
\begin{figure}[t]
    \centering
    \includegraphics[width=.48\textwidth]{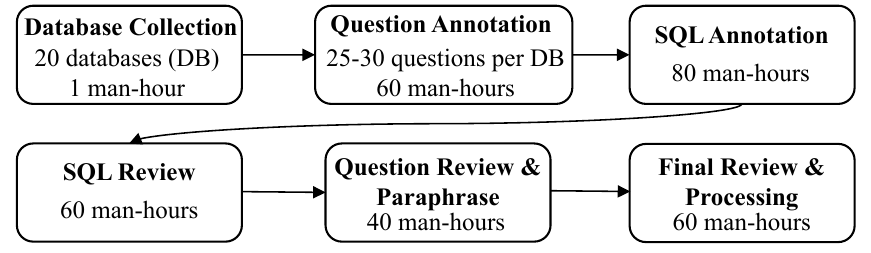}
    \caption{The annotation process of our Archer.}
    \label{fig:framework}
\end{figure}

As illustrated in Figure~\ref{fig:framework}, we create Archer in the following six steps, spending around 300 hours of human labor in total:  \S \ref{subsec:database collection} Database Collection, \S \ref{subsec:QRK} Question Annotation, \S \ref{subsec:sql anno} SQL Annotation, \S \ref{subsec:sql review} SQL review, \S \ref{subsec:para} Question Review and Paraphrase, \S \ref{subsec:final review} Final Review and Process.

\subsection{Database Collection}
\label{subsec:database collection}
%Relational databases have become widely adopted in both industry and academia. However, most of them are not publicly accessible.
In a noteworthy research study conducted by \citet{yu-etal-2018-spider}, a total of 200 high-quality databases across various domains were meticulously collected and created, requiring approximately 150 man-hours. Out of these, 166 databases were made publicly available.

%We obtained our databases from the 166 Spider databases. %Instead of utilizing all of these databases, we carefully selected a subset because 
Since not all Spider databases support the proposed reasoning types, 
%Our databases are selected based on
we carefully selected 20 databases across 20 domains from the Spider 166 databases based 
two criteria. Firstly, we applied a
script to keep only databases with a minimum of 3 tables and 20 columns within each database, as well as a minimum of 6 columns with time or numeric data types.
Secondly, we manually checked the filtered databases. %Secondly, we ensured that the selected databases included columns with either time or numeric data types.
%Specifically, we implemented a two-step filtering process on the Spider databases.
%Initially, we applied an automated script to filter the Spider databases. This 
%Following the automatic filtering, we manually reviewed the databases that passed the initial filtering step. This manual inspection 
These two steps ensure that each selected database contains sufficient information to support complex reasoning.

%Finally, we collected 20 databases across 20 domains. It is worth mentioning that we did not apply any further processing on these databases, as they had already been corrected and organized by \citet{yu-etal-2018-spider} in their work on Spider.

\subsection{Question Annotation}
\label{subsec:QRK}
Two bilingual (English and Chinese) Ph.D. students with SQL experience were assigned the task of generating questions based on 20 databases. The annotators were required to propose 25-30 questions for each database, ensuring that the questions met the following four requirements:

\paragraph{1) Arithmetic Reasoning:} Each question should incorporate arithmetic reasoning. The annotators were expected to include a minimum of five questions for each arithmetic reasoning type (addition, subtraction, multiplication, division).

\paragraph{2) Hypothetical Reasoning:} At least five questions should involve hypothetical reasoning. For each question using hypothetical reasoning, the annotators were also required to propose a corresponding factual question.

\paragraph{3) Commonsense Reasoning:} The annotators were encouraged to propose questions that involve commonsense reasoning. However, the number of questions with commonsense reasoning was not strictly limited. This flexibility acknowledged that not all databases support commonsense reasoning, and not all arithmetic calculations necessitate it.

\paragraph{4) Complex SQL Grammar:} The annotators were encouraged to propose questions that require the utilization of complex SQL grammar, such as GROUP BY, ORDER BY, and JOIN.

\paragraph{}The annotators were asked to write each question in both English and Chinese. Besides, they were instructed to indicate the reasoning types involved (arithmetic: addition, subtraction, multiplication, division; hypothetical; commonsense), and provide the relevant knowledge or formulation if the question incorporated commonsense reasoning. 

\subsection{SQL Annotation}
\label{subsec:sql anno}
In order to mitigate cognitive bias, we employed a diverse set of annotators for the tasks of generating questions and writing SQL queries. Two Ph.D. students, who possess strong SQL skills, were specifically chosen to translate the natural language questions into SQL queries. Their responsibilities encompassed the following: 
\paragraph{1) Clarity Ensuring:}The annotators reviewed both English and Chinese questions to identify any ambiguity and restructure them accordingly.
\paragraph{2) SQL Writing:}The annotators were instructed to use consistent SQL patterns when multiple equivalent queries are applicable for similar questions.
\paragraph{3) Verification and Correction:}The annotators were also responsible for reviewing the annotations pertaining to reasoning types and the common knowledge necessary to solve each question.

\subsection{SQL Review}
\label{subsec:sql review}
To ensure the correctness of the annotated SQL for each question, we employed a professional SQL expert to review all the SQL queries and rectify any incorrect ones. Subsequently, the original SQL annotators were responsible for verifying the SQL queries corrected by the expert. In cases where there are differences of opinion between the expert and the annotators regarding the corrected queries, they were required to engage in a discussion and reach a consensus to finalize the SQL annotation.

\subsection{Question Review and Paraphrase}
\label{subsec:para}
We employed two native English speakers and two native Chinese speakers to review and paraphrase English and Chinese questions, respectively. Initially, their task was to assess the naturalness and grammatical accuracy of the questions. Subsequently, the annotators were requested to provide a paraphrased version of each question in order to enhance the dataset's robustness.

\subsection{Final Review and Processing}
\label{subsec:final review}
In the final stage of our process, we assigned the task of reviewing the English and Chinese questions, SQL, and annotations relating to reasoning types and commonsense knowledge to our most seasoned annotator. Once this comprehensive review was completed, we ran a script to ensure that all SQL queries are executable.

\section{Dataset Statistics and Comparison}

\begin{table*}[t]
\centering
\resizebox{\textwidth}{!}{%
\begin{tabular}{c|cccccc|ccccccc|ccccccc|c}
\toprule
\multirow{2}{*}{\textbf{Dataset}} &
  \multicolumn{6}{c|}{\textbf{Scale}} &
  \multicolumn{7}{c|}{\textbf{Complexity}} &
  \multicolumn{7}{c|}{\textbf{Reasoning Distribution}} &
  \multirow{2}{*}{\textbf{Lang}} \\
    &
   \textbf{\#Q}&
   \textbf{\#SQL}&
   \textbf{\#DB}&
   \textbf{\#Dom}&
   \textbf{T/DB}&
   \textbf{C/DB}&
  \textbf{QL} &
  \textbf{SQLL} &
  \textbf{VS} &
  \textbf{TM} &
  \textbf{NL} &
  \textbf{GB} &
  \textbf{OB} &
  \textbf{A(+)} &
  \textbf{A(-)} &
  \textbf{A(*)}&
  \textbf{A(/)} &
  \textbf{H}&
  \textbf{C}&
  \textbf{C+H}&
   \\ \midrule
ATIS &
  5280 &
  947 &
  1 &
  1 &
  25 &
  131 &
  10.53 &
  \textbf{99.75} &
  3.14 &
  \textbf{4.66} &
  0.39 &
  0.01 &
  0.00 &
  \usym{2717} &
  \usym{2717} &
  \usym{2717} &
  \usym{2717} &
  \usym{2717} &
  \usym{2717} &
  \usym{2717} &
  en \\
GeoQuery &
  877 &
  246 &
  1 &
  1 &
  8 &
  31 &
  7.48 &
  26.76 &
  0.82 &
  1.46 &
  1.04 &
  0.18 &
  0.07 &
  \usym{2717} &
  \usym{2717} &
  \usym{2717} &
  0.2\% &
  \usym{2717} &
  \usym{2717} &
  \usym{2717} &
  en \\
Scholar &
  817 &
  193 &
  1 &
  1 &
  12 &
  28 &
  6.59 &
  38.03 &
  1.36 &
  3.26 &
  0.02 &
  0.37 &
  0.28 &
  \usym{2717} &
  0.5\% &
  \usym{2717} &
  \usym{2717} &
  \usym{2717} &
  \usym{2717} &
  \usym{2717} &
  en \\
Academic &
  196 &
  185 &
  1 &
  1 &
  15 &
  42 &
  13.33 &
  36.85 &
  1.30 &
  3.23 &
  0.04 &
  0.21 &
  0.12 &
  \usym{2717} &
  \usym{2717} &
  \usym{2717} &
  \usym{2717} &
  \usym{2717} &
  \usym{2717} &
  \usym{2717} &
  en \\
IMDB &
  131 &
  89 &
  1 &
  1 &
  16 &
  65 &
  10.23 &
  29.51 &
  1.20 &
  2.84 &
  0.01 &
  0.07 &
  0.11 &
  \usym{2717} &
  \usym{2717} &
  \usym{2717} &
  \usym{2717} &
  \usym{2717} &
  \usym{2717} &
  \usym{2717} &
  en \\
Yelp &
  128 &
  120 &
  1 &
  1 &
  7 &
  38 &
  9.87 &
  28.33 &
  1.68 &
  2.25 &
  0.00 &
  0.10 &
  0.08 &
  \usym{2717} &
  \usym{2717} &
  \usym{2717} &
  \usym{2717} &
  \usym{2717} &
  \usym{2717} &
  \usym{2717} &
  en \\
Advising &
  4387 &
  205 &
  1 &
  1 &
  18 &
  124 &
  10.90 &
  48.08 &
  3.06 &
  3.13 &
  0.17 &
  0.03 &
  0.07 &
  3.4\% &
  \usym{2717} &
  \usym{2717} &
  \usym{2717} &
  \usym{2717} &
  \usym{2717} &
  \usym{2717} &
  en \\
Restaurant &
  378 &
  23 &
  1 &
  1 &
  3 &
  12 &
  10.13 &
  29.57 &
  2.26 &
  2.26 &
  0.17 &
  0.00 &
  0.00 &
  \usym{2717} &
  \usym{2717} &
  \usym{2717} &
  \usym{2717} &
  \usym{2717} &
  \usym{2717} &
  \usym{2717} &
  en \\
WikiSQL &
  80654 &
  51159 &
  26531 &
  - &
  1 &
  6.33 &
  12.46 &
  13.32 &
  0.53 &
  1.00 &
  0.00 &
  0.00 &
  0.00 &
  \usym{2717} &
  \usym{2717} &
  \usym{2717} &
  \usym{2717} &
  \usym{2717} &
  \usym{2717} &
  \usym{2717} &
  en \\
DuSQL &
  25003 &
  20308 &
  208 &
  - &
  4.04 &
  21.38 &
  19.20 &
  20.63 &
  1.16 &
  1.33 &
  0.20 &
  0.42 &
  0.30 &
  2.4\% &
  9.5\% &
  1.0\% &
  4.4\% &
  \usym{2717} &
  - &
  \usym{2717} &
  zh \\
  BIRD &
   10962 &
   10841 &
   80 &
   -&
   7.68 &
   54.71 &
   15.81 &
   23.85 &
   1.16 &
   2.20 &
   0.08 &
   0.10 &
   0.19 &
   0.8\%&
   5.0\%&
   7.9\%&
   10.0\%&
  \usym{2717} &
  - &
  \usym{2717} &
  en \\
Cspider &
   9693 &
   5275 &
   166 &
   99 &
   5.28 &
   27.13 &
   11.90 &
   24.37 &
   0.93 &
   1.69 &
   0.10 &
   0.23 &
   0.21 &
   0.1\%&
   0.1\%&
   \usym{2717}&
   0.0\% &
   \usym{2717}&
   \usym{2717}&
   \usym{2717} &
  \underline{zh} \\
Spider &
  9693 &
  5275 &
  166 &
  99 &
  5.28 &
  27.13 &
  13.29 &
  24.37 &
  0.93 &
  1.69 &
  0.10 &
  0.23 &
  0.21 &
  0.1\% &
  0.1\% &
  \usym{2717} &
  0.0\% &
  \usym{2717} &
  \usym{2717} &
  \usym{2717} &
  en \\ 
  KaggleDBQA &
  272 &
  249 &
  8 &
  8 &
  2.13 &
  22.38 &
  9.83 &
  13.80 &
  0.54 &
  1.18 &
  0.00 &
  0.44 &
  \textbf{0.50} &
  0.0\% &
  0.0\% &
  \usym{2717} &
  0.0\% &
  \usym{2717} &
  \usym{2717} &
  \usym{2717} &
  en \\ 
  \midrule
\begin{tabular}[c]{@{}c@{}} \textbf{Archer} \emojiexample)\\ \textbf{(Ours)}\end{tabular} &
  1042 &
  521 &
  20 &
  20 &
  7.55 &
  45.25 &
  \begin{tabular}[c]{@{}c@{}}en-\textbf{29.94}\\ zh-\textbf{25.99}\end{tabular} &
  79.71&
  \textbf{6.21} &
  2.17 &
  \textbf{1.08} &
  \textbf{0.59} &
  0.26 &
  34.0\% &
  47.8\% &
  62.0\% &
  40.7\% &
  44.0\% &
  51.4\% &
  22.1\% &
  \begin{tabular}[c]{@{}c@{}}en\\ \underline{zh}\end{tabular} \\ \bottomrule
\end{tabular}%
}
\caption{Comparison of public text-to-SQL datasets. The abbreviations used are as follows: \#Q for the number of unique questions, \#SQL for the number of unique SQLs, \#DB for the number of databases, \#Dom for the number of domains, T/DB for the number of tables per database, C/DB for the number of columns per database, QL for the average question length, SQLL for the average SQL length, VS for the average number of value slots per question, TM for the average number of tables mentioned in each SQL, NL for the average nested level per SQL, GB and OB for the average number of \texttt{GROUP BY} and \texttt{ORDER BY} clauses per SQL respectively. A, H, C, and Lang represent arithmetic, hypothetical, commonsense, and language, respectively. The cross mark, - denote absence and presence respectively. The statistics for BIRD, CSpider, and Spider is based on training and dev sets as their test sets are unavailable. Language is represented as en for English databases and questions, zh for Chinese databases and questions, and \underline{zh} for English databases and Chinese questions.}
\label{tab:statistic}
\end{table*}

In Table~\ref{tab:statistic}, we present a summary of the statistics for Archer as well as other publicly available text-to-SQL datasets. We conducted a comparative analysis of Archer and other datasets based on four key perspectives: scale, complexity, reasoning distribution, and language. 

\subsection{Scale}  
Archer consists of 1,042 Chinese questions, 1,042 English questions, and 521 corresponding SQL queries, covering a wide range of 20 distinct databases spanning 20 domains. Each database in Archer, on average, consists of 7.55 tables and 45.25 columns.
Archer stands out for its inclusion of multiple domains and a higher average number of tables and columns. 

It is worth noting that WikiSQL~\cite{zhong2017seq2sql} and DuSQL~\cite{wang-etal-2020-dusql} are exceptionally large databases generated automatically. Inspired by them, Archer has the potential to serve as a valuable resource for summarizing SQL templates and training SQL-to-text generators to create large-scale datasets in line with our reasoning setting. In this project, we do not utilize Archer for automatic question-SQL pairs generation. This possibility is a potential future direction.

\subsection{Complexity}
Archer distinguishes itself by its considerably higher level of complexity compared to existing text-to-SQL datasets. Several factors contribute to this complexity:

Firstly, the average question length in Archer is significantly longer than that in other datasets. This poses a challenge to models because longer inputs increase the likelihood of misunderstandings or misinterpretations of specific question details.

Secondly, the average SQL length in Archer stands at 79.71, which is significantly longer than that of other datasets except for ATIS, which contains only one table. Longer SQL statements increase the likelihood of generating incorrect code.%, as errors or incorrect assumptions made early in the generation process can cascade and accumulate throughout the rest of the code.

Thirdly, value prediction, which is crucial in SQL generation, is often undervalued in current research. Interestingly, \citet{pourreza2023din} achieved an execution accuracy of 85.3\% on the Spider dataset without utilizing database content. This is primarily because Spider SQL queries typically contain an average of only 0.93 value slots, with most values explicitly quoted in the question. In contrast, Archer emphasizes the importance of values, with an average of 6.21 value slots per SQL. Furthermore, Archer questions do not explicitly quote exact values; instead, they naturally mention value information, mirroring real-world scenarios.

Fourthly, SQL queries in Archer refer to an average of 2.17 tables, suggesting that a substantial number of the questions require the use of information from multiple tables to derive SQLs.

Fifthly, the level of SQL statement nesting in Archer is higher than that in other datasets, indicating a greater degree of reasoning complexity required to answer Archer questions, which often necessitates the use of multiple subqueries.

Finally, Archer exhibits a high usage rate of complex SQL grammar features such as \texttt{GROUP BY} and \texttt{ORDER BY} in each SQL, surpassing the frequency of usage seen in nearly all other datasets.

\subsection{Reasoning Distribution}
All questions in Archer require arithmetic reasoning. This means that mathematical calculations and operations are essential in understanding and answering these questions effectively.
Additionally, 44.0\% of the questions involve hypothetical reasoning, where the model needs to reason about hypothetical scenarios to derive the correct SQL. Furthermore, 51.4\% of the questions require commonsense reasoning, where the model needs to utilize general knowledge and commonsense understanding to produce the correct SQL.

It is worth noting that the majority of previous text-to-SQL datasets do not incorporate arithmetic and commonsense reasoning.  Moreover, none of the previous datasets contain questions that involve hypothetical reasoning. Therefore, the inclusion of these types of reasoning tasks in Archer sets it apart from previous datasets and presents new challenges for models in the field of text-to-SQL understanding and generation.

\subsection{Language}
Unlike most previous text-to-SQL datasets that focus solely on English, Archer provides both English and Chinese   questions. %It is important to note that the databases in Archer are maintained in English as CSpider.
This bilingual feature of Archer enhances the evaluation and training capabilities of text-to-SQL models, catering to the needs of users in both English and Chinese languages, while forming a solid base for potential support of more languages for Archer, which is left as a future work.

\begin{table*}[t]\tiny
\centering
\resizebox{\textwidth}{!}{%
% \small
\begin{tabular}{lcccccccc}
\toprule
% \rule{0pt}{2ex} 
\multirow{3}{*}{\textbf{Models}} &
  \multicolumn{4}{c}{\rule{0pt}{2ex}\textbf{EN Questions, EN databases}} &
  \multicolumn{4}{c}{\rule{0pt}{2ex}\textbf{ZH Questions, EN databases}} \\
  [-.5ex]
  \cmidrule(lr){2-5} \cmidrule(lr){6-9}
 &
  \multicolumn{2}{c}{\textbf{Full}} &
  \multicolumn{2}{c}{\textbf{Test}} &
  \multicolumn{2}{c}{\textbf{Full}} &
  \multicolumn{2}{c}{\textbf{Test}} \\
  [-.5ex]
  \cmidrule(lr){2-3} \cmidrule(lr){4-5} \cmidrule(lr){6-7} \cmidrule(lr){8-9} 
 &
  \textbf{VA} &
  \textbf{EX} &
  \textbf{VA} &
  \textbf{EX} &
  \textbf{VA} &
  \textbf{EX} &
  \textbf{VA} &
  \textbf{EX} \\ \hline
\rowcolor{yellow} \multicolumn{9}{c}{\textit{\textbf{LLMs}}} \\ 
GPT-3.5 + API Doc &
  82.63 &
  13.24 &
  83.65 &
  3.85 &
  86.18 &
  10.65 &
  85.58 &
  \textbf{3.85} \\
GPT-3.5 + CT-3 &
  \textbf{84.17} &
  \textbf{13.34} &
  80.77 &
  3.85 &
  \textbf{91.17} &
  \textbf{12.86} &
  \textbf{91.35} &
  1.92 \\
GPT-3.5 + CT-3 + COT &
  75.14 &
  13.24 &
  74.04 &
  4.81 &
  72.84 &
  12.19&
  65.38 &
  \textbf{3.85}\\
GPT-4 + DIN-SQL &
  - &
  - &
   \textbf{96.15}&
   \textbf{6.73}&
  - &
  - &
  - &
  - \\ \hline
\rowcolor{yellow}\multicolumn{9}{c}{\textit{\textbf{Fine-tuned Models}}} \\ 
T5-base/mT5-base &
  - &
  - &
  11.54 &
  0.00 &
  - &
  - &
  \multicolumn{1}{r}{9.62} &
  0.00 \\
T5-large/mT5-large &
  - &
  - &
  15.38 &
  0.00 &
  - &
  - &
  \multicolumn{1}{r}{14.42} &
  0.00 \\
T5-3B/mT5-xl &
  - &
  - &
  19.23 &
  0.00 &
  - &
  - &
  \multicolumn{1}{r}{17.31} &
  0.00 \\ 
T5-base/mT5-base + Aug &
  - &
  - &
  25.00&
  0.00&
  - &
  - &
  24.03 &
  0.00\\
T5-large/mT5-large + Aug  &
  - &
  - &
  33.65 &
  3.84&
  - &
  - &
  30.77 &
  0.96\\
T5-3B/mT5-xl + Aug &
  - &
  - &
  \textbf{50.00} &
  \textbf{4.81} &
  - &
  - &
  \textbf{61.54} &
  \textbf{1.92} \\\bottomrule
\end{tabular}%
}
\caption{Baseline performance on Archer. GPT-4+DIN-SQL was tested only on the English set due to cost and its English-specific design. We only report the fine-tuned model's performance on the test set.\vspace{-.2cm}}
\label{tab:overall performance}
\end{table*}

\section{Experiments}
\subsection{Baseline Models}
We benchmark the performance of two types of presentative text-to-SQL models on Archer: LLMs and finetuned Models.

\paragraph{LLMs} LLMs have shown strong performance on commonly used text-to-SQL benchmarks, such as Spider. 
To analyze the difficulty of the whole Archer, we provide the zero-shot results of GPT-3.5 (\texttt{\href{https://platform.openai.com/docs/models/gpt-3-5}{gpt-3.5-turbo}}) with different prompt settings: \textit{API Doc}, \textit{CT-3}, \textit{CT-3+COT}.
\textit{API Doc} follows the style of the Text-to-SQL example provided by OpenAI, which includes the schema information in a comment style. 
\textit{CT-3}, introduced by \citet{rajkumar2022evaluating}, includes the
\texttt{CREATE TABLE} commands for each table and the results of executing a \texttt{SELECT * FROM T LIMIT 3} query on each table. Compared with API Doc, CT-3 provides more information like declarations of column types and foreign keys, and a small amount (3) of content examples.
\textit{CT-3+COT} implement the Chain-Of-Thought (COT) technique on top of the CT-3 prompt by appending the prompt sentence \texttt{"Let’s think step by step."} before the SQL generation. Following the work of \citet{li2023can}, we provide a 1-shot pseudo example for LLMs to learn the procedure of thinking and output format.
Furthermore, we evaluate the performance of \textit{GPT-4+DIN-SQL} \cite{pourreza2023din} on Archer. As a highly-ranked solution on the Spider leaderboard at the time of writing, it consists of four modules: (1) schema linking, (2) query classification and decomposition, (3) SQL generation, and (4) self-correction. The initial three modules exploit the in-context learning ability of GPT-4 with ten shots, while the self-correction is conducted by GPT-4 in a zero-shot setting. Note that we do not evaluate GPT-4+DIN-SQL on Archer Chinese questions because it is designed for English datasets.
More details on the prompts can be found in Appendix~\ref{App: prompts}.

\paragraph{Fine-tuned Models} T5-based fine-tuned models have shown promising results on the Spider leaderboard. It is, however, worth mentioning that many top-tier models on the leaderboard are customized specifically for the limited SQL grammars present in the Spider dataset. Given that our dataset contains more complex grammatical structures compared to Spider, these specialized models may not be suitable for our needs. As a result, we select vanilla T5 models as our baselines instead of the aforementioned variants. We evaluate English questions using
\href{https://huggingface.co/t5-base}{T5-base}, \href{https://huggingface.co/t5-large}{T5-large}, \href{https://huggingface.co/t5-3b}{T5-3B}, and evaluate Chinese questions using  \href{https://huggingface.co/google/mt5-base}{mT5-base}, \href{https://huggingface.co/google/mt5-large}{mT5-large}, \href{https://huggingface.co/google/mt5-xl}{mT5-xl}.
We concatenate the natural question \textit{Q} and database schema into a sequence as input in a format as below:
\begin{equation}
\small
x=[q_1,...,q_{|Q|}|t_1\mathbin{:} c_1^{t_1},...,c_{|t_1|}^{t_1}|...|t_{|\mathcal{T}|}\mathbin{:}  c_{1}^{t_{|\mathcal{T}|}},...,c_{|t_{|\mathcal{T}|}|}^{t_{|\mathcal{T}|}}]
\end{equation}
where $q_i$ is the $i^{th}$ question token, $t_j$ is the $j^{th}$ table, and $c_k^{t_j}$ is the $k^{th}$ in  the $j^{th}$ table.
Following the works of \citet{li2023resdsql,lin2020bridging}, we extract the potential database cell values and append them to their corresponding columns.

\subsection{Evaluation Metrics} 
We employ two evaluation metrics: VAlid SQL (VA) and EXecution accuracy (EX).
VA is the proportion of the predicted SQL statements that can be executed successfully, no matter  with correct or incorrect answers.
EX is the proportion of the predicted SQL statements where the execution results match those of  the gold SQL statements. We computed EX of each instance use a new evaluation script as shown in Algorithm~\ref{algorithm}, which mitigates the false-negative issue in present publicly available evaluation scripts caused by permutations of columns and rows.

\begin{figure*}[t]
    \centering
    \includegraphics[width=\textwidth]{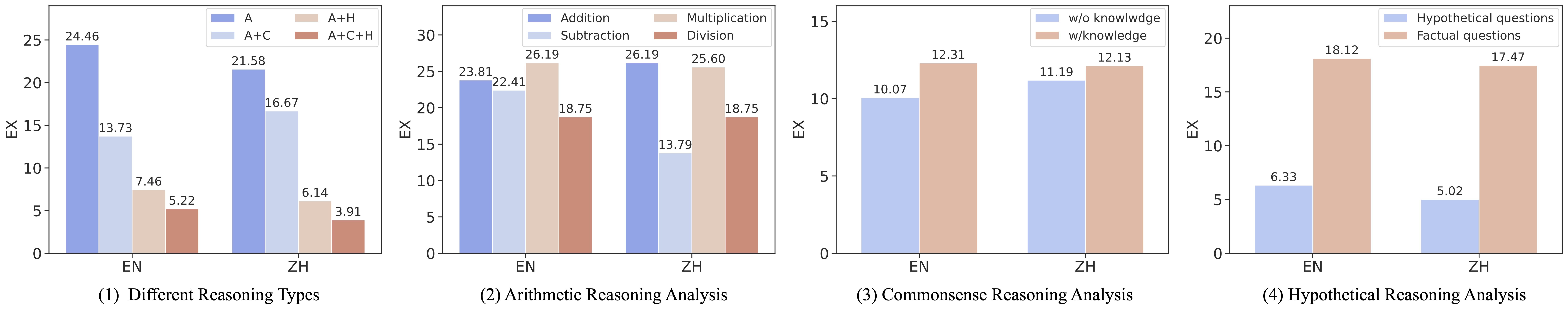}
    \caption{GPT-3.5 + CT-3 execution accuracy comparison across and within different reasoning types. A refers to arithmetic. H refers to hypothetic. C refers to commonsense. \vspace{-.2cm}}
    \label{fig:reason_all}
\end{figure*}

\subsection{Experiments Setup} 
\paragraph{Data Split} Among the %consists of 
20 databases, we split 16, 2, and 2 databases as training, dev, and test sets, respectively.
%It should be noted that 
The databases for Archer training set are collected from the Spider training set, and the databases for Archer dev set and test set are collected from the Spider dev set. We strive to introduce as few new SQL keywords as possible during SQL annotation to facilitate the integration of our dataset with the Spider and CSpider datasets. 
We also report the performance of T5 finetuned on the augmented training set which consists of Archer training set and Spider/CSpider training set.
For LLM baselines, we assess the zero-shot performance of GPT-3.5 on the full Archer to evaluate the dataset's overall difficulty. As for GPT-4+DINSQL, due to its high cost and extended response time, we only test it on Archer test sets.

\paragraph{Hyper-Parameters} For GPT-3.5 baselines, we set stop sequence to \texttt{[`-\,-', `;', `\#']} and the temperature to 0. In the case of GPT-4+DIN-SQL, we adhere to the default setting as outlined in \citet{pourreza2023din}. For T5 baselines, we employ the Adafactor optimizer with a learning rate of 5e-5. For T5-base/mT5-base and T5-large/mT5-large, we adopt a batch size of 6 and a gradient descent step of 5. For T5-3b and mT5-xl, we use a batch size of 2 and a gradient descent step of 16.
To adjust the learning rate, we utilize linear warm-up with a warm-up rate of 0.1, followed by cosine decay.
During inference, we set the beam size to 8. We set the maximum epoch to 128, having checkpoints every 10 epochs as well as the last epoch. We then select the optimal checkpoints based on their EX performance on the development set.

\section{Results and Discussion}
\subsection{Overall Evaluation}
We summarize the performance of LLMs and finetuned models in Table~\ref{tab:overall performance}. The low performance of these models on Archer suggests that Archer %our text-to-SQL dataset %, requiring complex reasoning, 
presents a significant challenge. This underscores the considerable potential for future improvement in this domain.

\paragraph{LLM} GPT-4+DIN-SQL obtain EX score of 6.73\% on Archer test set, while it is able to achieve 85.3\% test-suite execution performance on Spider test set ~\cite{pourreza2023din}. 
To evaluate the overall difficulty of Archer, we test the zero-shot performance of GPT-3.5 with API Doc, CT-3, CT-3+COT prompts on the full Archer data. Among the three kinds of prompts, CT-3 achieves the highest EX scores on both English data (EX: 13.34\%) and Chinese data (EX: 12.86\%). As expected, CT-3 performed slightly better than API Doc, likely due to its inclusion of more useful information, such as declarations of column types and foreign keys. However, the addition of COT in CT-3+COT did not outperform CT-3 on the complete Archer. On the other hand, for the Test set only, CT-3+COT slightly outperform CT-3.
From Table~\ref{tab:overall performance}, we observe a significant decrease in VA when using COT,  suggesting that %while COT has the potential to enhance SQL generation for complex reasoning, it may introduce 
COT suffers from having more syntax errors in the generated SQL. % due to the language-SQL barrier. 
%Furthermore, although COT excels in solving math questions, Archer encompasses a substantial proportion of questions that require other reasoning abilities.
Although CT-3+COT achieved a higher EX score than CT-3 and API Doc specifically for questions involving arithmetic and commonsense reasoning, % However, 
it performed less effectively on questions that require hypothetical reasoning (cf. Table~\ref{tab:performance wrt reasoning-full} in Appendix~\ref{App:rea-results}). 

\paragraph{Finetuned Models} From Table~\ref{tab:overall performance}. we observe that T5 from scale base to 3B (XL) trained on Archer training set achieve 0.00\% EX scores. This outcome could be attributed to the small-scale nature of Archer combined with its high complexity.
However, when Archer training set was augmented with the Spider/CSpider training set, the VA scores of T5 models exhibited a substantial improvement. Specifically, the T5-3B model trained on the augmented training set achieved an EX score of 4.81\% on the English test (matching the performance of GPT-3.5+CT-3+COT) set and 1.92\% on the Chinese test set (matching the performance of GPT-3.5+CT-3). 

These results suggest that  Archer has the potential to advance the development of text-to-SQL systems with complex reasoning. %However, further exploration and investigation in this area remain as future work.

\subsection{Different Reasoning Analysis}
To gain a comprehensive understanding of the difficulty levels within the complete Archer across various reasoning types, we conducted a thorough analysis using the GPT-3.5 model with the CT-3 prompt, which demonstrated the highest performance on the full dataset. Additional results for GPT-3.5 with alternative prompts can be found in  Appendix~\ref{App:rea-results}.

\paragraph{Overall Comparison}Figure~\ref{fig:reason_all}-(1) shows the performance on questions with different kinds of reasoning.  The results reveal that questions solely based on arithmetic reasoning exhibit significantly higher performance compared to those involving additional forms of reasoning. Specifically, hypothetical reasoning presents a greater challenge than commonsense reasoning. Moreover, questions that require the integration of all three reasoning types exhibit the poorest performance.

\paragraph{Arithmetic Reasoning}The performance on questions that exclusively require arithmetic reasoning across various arithmetic operations is presented in Figure~\ref{fig:reason_all}-(2). The findings indicate that subtraction and division pose greater difficulty compared to addition and multiplication.

\paragraph{Commonsense Reasoning} %Archer provides additional annotations regarding inferred knowledge for each question involving 
On commonsense reasoning, in Figure~\ref{fig:reason_all}-(3), we compare the performance of GPT-3.5+CT-3 on such questions under two settings. The first setting involves directly inputting the question itself, while the second setting involves inputting the concatenation of the knowledge and the question. The results reveal that explicitly stating the knowledge within the question can aid in generating correct SQL queries.
This 
%observation 
suggests that leveraging external knowledge bases could be beneficial in solving similar questions. However, incorporating external knowledge into text-to-SQL tasks presents significant challenges in general. Firstly,   models need to compare information from natural language questions with the relational database to determine if external knowledge is required. Secondly,   models need to extract the most relevant knowledge from external knowledge bases. Last but not least, the process of integrating this knowledge into the text-to-SQL generation process remains largely unexplored.

\paragraph{Hypothetical reasoning} %In Archer, every
On hypothetical reasoning, % is paired with a corresponding factual question. In 
in Figure~\ref{fig:reason_all}-(4), we compare the performance of these questions and observe a significant performance gap. The EX performance on factual questions exceeds 17\%, whereas the performance on hypothetical questions falls below 7\%, confirming %. This indicates that the models encounter 
the difficulties involved. %when confronted with counterfactual information in the questions.

\subsection{Complexity Factors Analysis}
\begin{figure}[t]
    \centering
    \includegraphics[width=.45\textwidth]{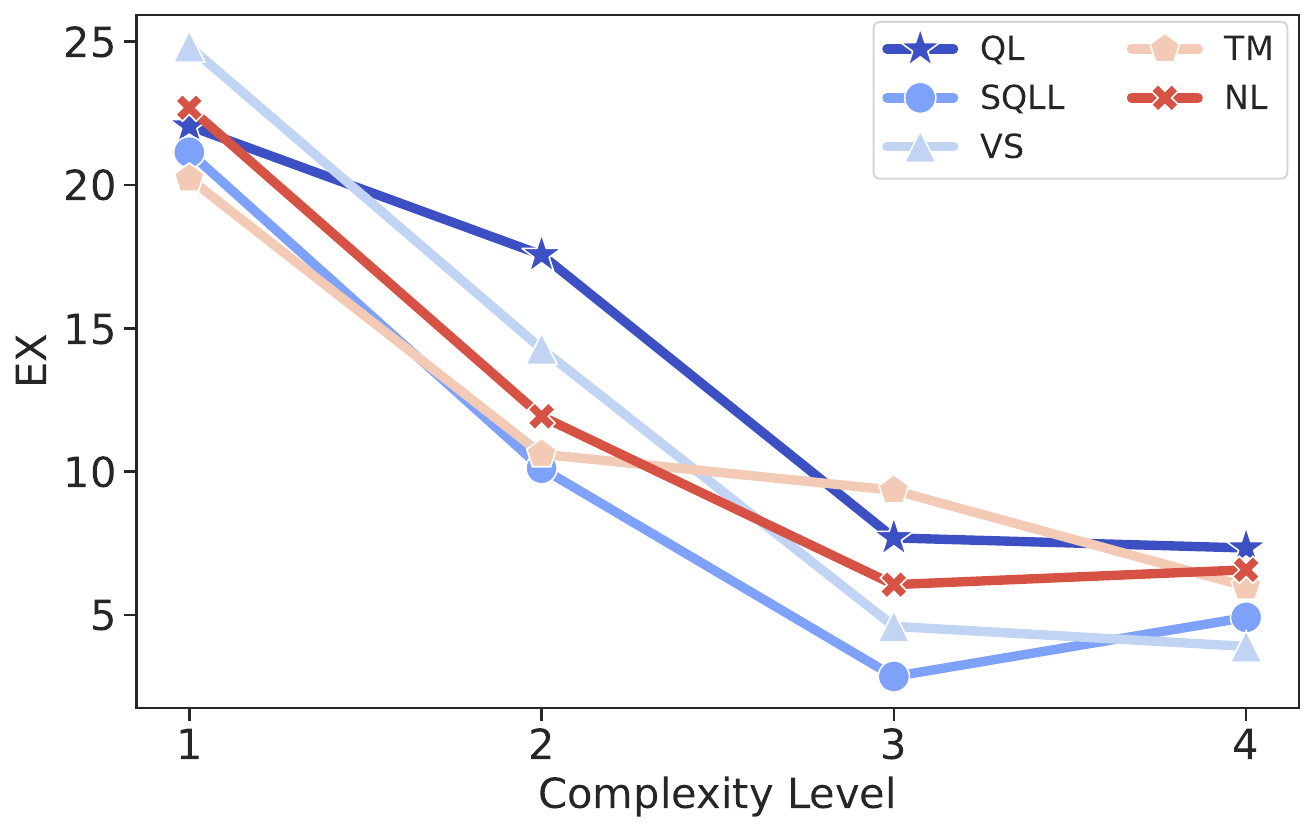}
    \caption{GPT-3.5 + CT-3 execution accuracy performance w.r.t different complexity level. The abbreviations used are as follows: QL for the average question length (1: [0,15)], 2: [15,20), 3:[30,45), 4: [45,)), SQLL for the average SQL length (1: [0,50)], 2: [50,100), 3:[100,150), 4: [150,)), VS for the average number of value slots per question (1: [0,3)], 2: [3,6), 3:[6,9), 4: [9,)), TM for the average number of tables mentioned in each SQL (1: [0,2)], 2: [2,3), 3:[3,5), 4: [5,)), NL for the average nested level per SQL (1: [0,1)], 2: [1,2), 3:[2,3), 4: [3,)).}
    \label{fig:complextity}
\end{figure}

To gain insights into the SQL complexity within Archer, Figure~\ref{fig:complextity} illustrates the relationship between the EX score and various factors, including question length, SQL length, number of value slots, number of tables mentioned in SQL, and SQL nested level.  The performance demonstrates a decreasing trend as the question becomes longer, the SQL length increases, the number of value slots rises, the number of tables mentioned in the SQL grows, or the SQL nested level escalates. 
As shown in Table~\ref{tab:statistic}, Archer exhibits considerably higher complexity across these factors when compared to other publicly available text-to-SQL datasets.

\subsection{Bad Case Analysis}
We randomly selected 50 executable but incorrect examples generated by GPT-4 + DIN-SQL and identified the following common error types:
% \begin{enumerate}
 
\paragraph{Incorrect Logic}: GPT-4 sometimes struggles with hypothetical questions that involve complex logic. For instance, when asked "If all cars produced by Daimler Benz company are 4-cylinders, which 4-cylinder car needs the most fuel to drive 300 miles?", the model might generate SQL queries like \texttt{WHERE T1.Cylinders = 4 AND T4.Maker = 'Daimler Benz'}. However, the correct query should be \texttt{WHERE T1.Cylinders = 4 OR T4.Maker = `Daimler Benz'} as there could be other 4-cylinder cars aside from Mercedes-Benz. This reveals a limitation in comprehending the hypothetical nature of the question.

\paragraph{Incorrect Knowledge}: GPT-4 may make commonsense errors when generating the SQL, such like unit conversions. For example, if a question requests fuel consumption in liters per hundred kilometers, but the database only contains fuel efficiency data in miles per gallon, the accurate conversion formula is \texttt{liters\_per\_hundred\_kilometers = 235.2145 / MPG}. However, GPT-4 employs an incorrect formula like \texttt{(100 * 3.78541) / MPG}.

\paragraph{Incorrect Schema Understanding}: GPT-4 sometimes struggles to correctly link query entities to the corresponding database columns. For example, when asked about the "average single cylinder displacement of an 8-cylinder car", GPT-4 might generate a query like \texttt{SELECT avg(Edispl) FROM cars\_data WHERE Cylinders = 8}. However, in this case, the query should calculate the average single cylinder displacement, like \texttt{SELECT AVG(1.0 * Edispl / Cylinders) AS avg\_displ FROM cars\_data WHERE Cylinders = 8}. This error highlights the need for the model to understand database column names, especially when they involve abbreviations commonly used in real-world databases. (Note that in the Spider dataset, annotators tend to use exact column names in their queries, e.g., What is the average edispl for all Volvos?)

\paragraph{Other Detail Errors}: For example, GPT-4 may also exhibit minor errors such as forgetting to multiply 1.0 for float calculations.
% \end{enumerate}

\section{Related Work}
The earliest text-to-SQL datasets, including ATIS~\cite{dahl-etal-1994-expanding, iyer2017learning}, GeoQuery~\cite{zelle1996learning,iyer2017learning}, Scholar~\cite{iyer2017learning}, Academic~\cite{data-academic}, IMDB~\cite{data-sql-imdb-yelp}, Yelp~\cite{data-sql-imdb-yelp}, Advising~\cite{data-sql-advising} and Restaurants~\cite{data-restaurants,data-restaurants-logic,data-restaurants-original}, were limited to a single database.
Consequently, models trained on these datasets struggled to generalize to unseen databases as they were tested on the same database used for training.
To address such limitations, \citet{zhong2017seq2sql} introduced WikiSQL in which the databases in the test set were not present in the training set. However, the SQL queries in WikiSQL were generated automatically using simplified assumptions, which may not fully capture the complexity of real-world queries.

For a comprehensive cross-domain text-to-SQL dataset, \citet{yu-etal-2018-spider} presented Spider dataset, which is currently the most widely used text-to-SQL datasets. 
However, Spider excludes questions that require external knowledge, like commonsense reasoning and mathematical calculations, which are often essential for real-world applications.

Then, \citet{wang-etal-2020-dusql} proposed DuSQL, a Chinese cross-domain text-to-SQL dataset that includes math-related questions. However, DuSQL's queries and questions are relatively simple due to automatic generation and grammar restrictions.
\citet{dou-etal-2022-towards} extended DuSQL with external knowledge in their KnowSQL dataset.
Unfortunately, KnowSQL is not publicly unavailable.

In real-life scenarios, databases can be dirtier with abbreviated and obscure naming of tables, columns, and data values. To address this, \citet{lee-2021-kaggle-dbqa} proposed KaggleDBQA with realistic databases.
\citet{li2023can} proposed BIRD benchmark for the text-to-SQL task on big and dirty databases with a total size of 33.4 GB.

In contrast to these existing text-to-SQL datasets, Archer focuses specifically on questions involving complex reasoning %, including arithmetic, commonsense, and hypothetical reasoning. Archer
and offers both English and Chinese questions to query English databases across various domains. Notably, all questions and SQL queries in Archer are manually annotated by humans and thoroughly reviewed by professionals, ensuring high-quality annotations for training and evaluation purposes.

\section{Conclusion}
In this paper, we present Archer, a complex bilingual text-to-SQL dataset with three distinct reasoning types: arithmetic, commonsense, and hypothetical reasoning. Experimental results on Archer, obtained from both LLMs and fine-tuned models, suggest plenty of space for improvement. 

\section*{Acknowledgement}
This work is supported by   Huawei’s Dean's Funding (C-00006589) and the UKRI Centre for Doctoral Training in Natural Language Processing, funded by the UKRI (grant EP/S022481/1).

\section*{Limitations}
% The primary limitation of our study arises from the relatively small scale of Archer, especially when compared to other cross-domain text-to-SQL datasets. The intricate reasoning needed for annotating such a dataset makes it both costly and challenging, hence restricting the dataset size. This, in turn, hampers the capacity for training finetuned models and potentially limits their overall performance. We strongly advocate for future research to (1) focus on developing models that can effectively work with smaller yet more complex datasets, and (2) utilize Archer as a high-quality foundational resource to generate additional question-SQL pairs, thereby enhancing the dataset's scope.
The evaluation metric used in Archer is execution accuracy. This metric may be perceived as an upper-bound performance measure, as SQL queries producing the same execution results on a single database may still possess different semantic meanings. To overcome this limitation, we plan to release a test suite in the future that evaluates SQL queries on multiple databases, allowing for a more comprehensive assessment of semantic accuracy.

\section*{Ethics Statement}
As mentioned in the submission, we select our databases from Spider~\cite{yu-etal-2018-spider}, which is public for academic use and does not contain sensitive information. 
The construction of our dataset involved the active involvement of human participants. We recruited and provided training to five annotators who possessed backgrounds in databases. These annotators were assigned the tasks of generating questions based on the databases, writing SQL queries, and paraphrasing the questions. Importantly, no sensitive personal information was involved throughout this process.
Our human annotation study underwent evaluation by the departmental ethics panel, which deemed it exempt from ethical approval. This exemption was based on the fact that all participants were employees of the University of Edinburgh and were therefore protected by employment law. Furthermore, participants received compensation at the standard hourly rate designated for tutors and demonstrators at the university.
To promote academic usage, we intend to freely release the dataset online.

% Entries for the entire Anthology, followed by custom entries
\bibliography{anthology,custom}

\begin{thebibliography}{21}
\expandafter\ifx\csname natexlab\endcsname\relax\def\natexlab#1{#1}\fi

\bibitem[{Dahl et~al.(1994)Dahl, Bates, Brown, Fisher, Hunicke-Smith, Pallett, Pao, Rudnicky, and Shriberg}]{dahl-etal-1994-expanding}
Deborah~A. Dahl, Madeleine Bates, Michael Brown, William Fisher, Kate Hunicke-Smith, David Pallett, Christine Pao, Alexander Rudnicky, and Elizabeth Shriberg. 1994.
\newblock \href {https://aclanthology.org/H94-1010} {Expanding the scope of the {ATIS} task: The {ATIS}-3 corpus}.
\newblock In \emph{{H}uman {L}anguage {T}echnology: Proceedings of a Workshop held at {P}lainsboro, {N}ew {J}ersey, {M}arch 8-11, 1994}.

\bibitem[{Dou et~al.(2022)Dou, Gao, Liu, Pan, Wang, Che, Zhan, Kan, and Lou}]{dou-etal-2022-towards}
Longxu Dou, Yan Gao, Xuqi Liu, Mingyang Pan, Dingzirui Wang, Wanxiang Che, Dechen Zhan, Min-Yen Kan, and Jian-Guang Lou. 2022.
\newblock \href {https://doi.org/10.18653/v1/2022.emnlp-main.350} {Towards knowledge-intensive text-to-{SQL} semantic parsing with formulaic knowledge}.
\newblock In \emph{Proceedings of the 2022 Conference on Empirical Methods in Natural Language Processing}, pages 5240--5253, Abu Dhabi, United Arab Emirates. Association for Computational Linguistics.

\bibitem[{Finegan-Dollak et~al.(2018)Finegan-Dollak, Jonathan K.~Kummerfeld, Ramanathan, Sadasivam, Zhang, and Radev}]{data-sql-advising}
Catherine Finegan-Dollak, Li~Zhang Jonathan K.~Kummerfeld, Karthik Ramanathan, Sesh Sadasivam, Rui Zhang, and Dragomir Radev. 2018.
\newblock \href {http://aclweb.org/anthology/P18-1033} {Improving text-to-sql evaluation methodology}.
\newblock In \emph{Proceedings of the 56th Annual Meeting of the Association for Computational Linguistics (Volume 1: Long Papers)}, pages 351--360.

\bibitem[{Giordani and Moschitti(2012)}]{data-restaurants}
Alessandra Giordani and Alessandro Moschitti. 2012.
\newblock \href {https://doi.org/10.1007/978-3-642-45260-4_5} {Automatic generation and reranking of sql-derived answers to nl questions}.
\newblock In \emph{Proceedings of the Second International Conference on Trustworthy Eternal Systems via Evolving Software, Data and Knowledge}, pages 59--76.

\bibitem[{Iyer et~al.(2017)Iyer, Konstas, Cheung, Krishnamurthy, and Zettlemoyer}]{iyer2017learning}
Srinivasan Iyer, Ioannis Konstas, Alvin Cheung, Jayant Krishnamurthy, and Luke Zettlemoyer. 2017.
\newblock Learning a neural semantic parser from user feedback.
\newblock \emph{arXiv preprint arXiv:1704.08760}.

\bibitem[{Lee et~al.(2021)Lee, Polozov, and Richardson}]{lee-2021-kaggle-dbqa}
Chia-Hsuan Lee, Oleksandr Polozov, and Matthew Richardson. 2021.
\newblock \href {https://aclanthology.org/2021.acl-long.176} {{KaggleDBQA}: Realistic evaluation of text-to-{SQL} parsers}.
\newblock In \emph{Proceedings of the 59th Annual Meeting of the Association for Computational Linguistics and the 11th International Joint Conference on Natural Language Processing (Volume 1: Long Papers)}, pages 2261--2273, Online. Association for Computational Linguistics.

\bibitem[{Li and Jagadish(2014)}]{data-academic}
Fei Li and H.~V. Jagadish. 2014.
\newblock \href {http://dx.doi.org/10.14778/2735461.2735468} {Constructing an interactive natural language interface for relational databases}.
\newblock \emph{Proceedings of the VLDB Endowment}, 8(1):73--84.

\bibitem[{Li et~al.(2023{\natexlab{a}})Li, Zhang, Li, and Chen}]{li2023resdsql}
Haoyang Li, Jing Zhang, Cuiping Li, and Hong Chen. 2023{\natexlab{a}}.
\newblock Resdsql: Decoupling schema linking and skeleton parsing for text-to-sql.
\newblock \emph{arXiv preprint arXiv:2302.05965}.

\bibitem[{Li et~al.(2023{\natexlab{b}})Li, Hui, Cheng, Qin, Ma, Huo, Huang, Du, Si, and Li}]{li2023graphix}
Jinyang Li, Binyuan Hui, Reynold Cheng, Bowen Qin, Chenhao Ma, Nan Huo, Fei Huang, Wenyu Du, Luo Si, and Yongbin Li. 2023{\natexlab{b}}.
\newblock Graphix-t5: Mixing pre-trained transformers with graph-aware layers for text-to-sql parsing.
\newblock \emph{arXiv preprint arXiv:2301.07507}.

\bibitem[{Li et~al.(2023{\natexlab{c}})Li, Hui, Qu, Li, Yang, Li, Wang, Qin, Cao, Geng et~al.}]{li2023can}
Jinyang Li, Binyuan Hui, Ge~Qu, Binhua Li, Jiaxi Yang, Bowen Li, Bailin Wang, Bowen Qin, Rongyu Cao, Ruiying Geng, et~al. 2023{\natexlab{c}}.
\newblock Can llm already serve as a database interface? a big bench for large-scale database grounded text-to-sqls.
\newblock \emph{arXiv preprint arXiv:2305.03111}.

\bibitem[{Lin et~al.(2020)Lin, Socher, and Xiong}]{lin2020bridging}
Xi~Victoria Lin, Richard Socher, and Caiming Xiong. 2020.
\newblock Bridging textual and tabular data for cross-domain text-to-sql semantic parsing.
\newblock In \emph{Findings of the Association for Computational Linguistics: EMNLP 2020}, pages 4870--4888.

\bibitem[{Popescu et~al.(2003)Popescu, Etzioni, and Kautz}]{data-restaurants-original}
Ana-Maria Popescu, Oren Etzioni, and Henry Kautz. 2003.
\newblock \href {http://doi.acm.org/10.1145/604045.604070} {Towards a theory of natural language interfaces to databases}.
\newblock In \emph{Proceedings of the 8th International Conference on Intelligent User Interfaces}, pages 149--157.

\bibitem[{Pourreza and Rafiei(2024)}]{pourreza2023din}
Mohammadreza Pourreza and Davood Rafiei. 2024.
\newblock Din-sql: Decomposed in-context learning of text-to-sql with self-correction.
\newblock In \emph{Proc. of NeurIPS 2024}.

\bibitem[{Rajkumar et~al.(2022)Rajkumar, Li, and Bahdanau}]{rajkumar2022evaluating}
Nitarshan Rajkumar, Raymond Li, and Dzmitry Bahdanau. 2022.
\newblock Evaluating the text-to-sql capabilities of large language models.
\newblock \emph{arXiv preprint arXiv:2204.00498}.

\bibitem[{Scholak et~al.(2021)Scholak, Schucher, and Bahdanau}]{scholak-etal-2021-picard}
Torsten Scholak, Nathan Schucher, and Dzmitry Bahdanau. 2021.
\newblock \href {https://doi.org/10.18653/v1/2021.emnlp-main.779} {{PICARD}: Parsing incrementally for constrained auto-regressive decoding from language models}.
\newblock In \emph{Proceedings of the 2021 Conference on Empirical Methods in Natural Language Processing}, pages 9895--9901, Online and Punta Cana, Dominican Republic. Association for Computational Linguistics.

\bibitem[{Tang and Mooney(2000)}]{data-restaurants-logic}
Lappoon~R. Tang and Raymond~J. Mooney. 2000.
\newblock \href {http://www.aclweb.org/anthology/W00-1317} {Automated construction of database interfaces: Intergrating statistical and relational learning for semantic parsing}.
\newblock In \emph{2000 Joint SIGDAT Conference on Empirical Methods in Natural Language Processing and Very Large Corpora}, pages 133--141.

\bibitem[{Wang et~al.(2020)Wang, Zhang, Wu, Sun, Li, Wu, Zhang, and Wang}]{wang-etal-2020-dusql}
Lijie Wang, Ao~Zhang, Kun Wu, Ke~Sun, Zhenghua Li, Hua Wu, Min Zhang, and Haifeng Wang. 2020.
\newblock \href {https://doi.org/10.18653/v1/2020.emnlp-main.562} {{D}u{SQL}: A large-scale and pragmatic {C}hinese text-to-{SQL} dataset}.
\newblock In \emph{Proceedings of the 2020 Conference on Empirical Methods in Natural Language Processing (EMNLP)}, pages 6923--6935, Online. Association for Computational Linguistics.

\bibitem[{Yaghmazadeh et~al.(2017)Yaghmazadeh, Wang, Dillig, , and Dillig}]{data-sql-imdb-yelp}
Navid Yaghmazadeh, Yuepeng Wang, Isil Dillig, , and Thomas Dillig. 2017.
\newblock \href {http://doi.org/10.1145/3133887} {Sqlizer: Query synthesis from natural language}.
\newblock In \emph{International Conference on Object-Oriented Programming, Systems, Languages, and Applications, ACM}, pages 63:1--63:26.

\bibitem[{Yu et~al.(2018)Yu, Zhang, Yang, Yasunaga, Wang, Li, Ma, Li, Yao, Roman, Zhang, and Radev}]{yu-etal-2018-spider}
Tao Yu, Rui Zhang, Kai Yang, Michihiro Yasunaga, Dongxu Wang, Zifan Li, James Ma, Irene Li, Qingning Yao, Shanelle Roman, Zilin Zhang, and Dragomir Radev. 2018.
\newblock \href {https://doi.org/10.18653/v1/D18-1425} {{S}pider: A large-scale human-labeled dataset for complex and cross-domain semantic parsing and text-to-{SQL} task}.
\newblock In \emph{Proceedings of the 2018 Conference on Empirical Methods in Natural Language Processing}, pages 3911--3921, Brussels, Belgium. Association for Computational Linguistics.

\bibitem[{Zelle and Mooney(1996)}]{zelle1996learning}
John~M Zelle and Raymond~J Mooney. 1996.
\newblock Learning to parse database queries using inductive logic programming.
\newblock In \emph{Proceedings of the national conference on artificial intelligence}, pages 1050--1055.

\bibitem[{Zhong et~al.(2017)Zhong, Xiong, and Socher}]{zhong2017seq2sql}
Victor Zhong, Caiming Xiong, and Richard Socher. 2017.
\newblock Seq2sql: Generating structured queries from natural language using reinforcement learning.
\newblock \emph{arXiv preprint arXiv:1709.00103}.

\end{thebibliography}

\appendix

\onecolumn
\section{Prompts}
\label{App: prompts}

\begin{figure}[H]
    \centering
    \includegraphics[width=.85\textwidth]{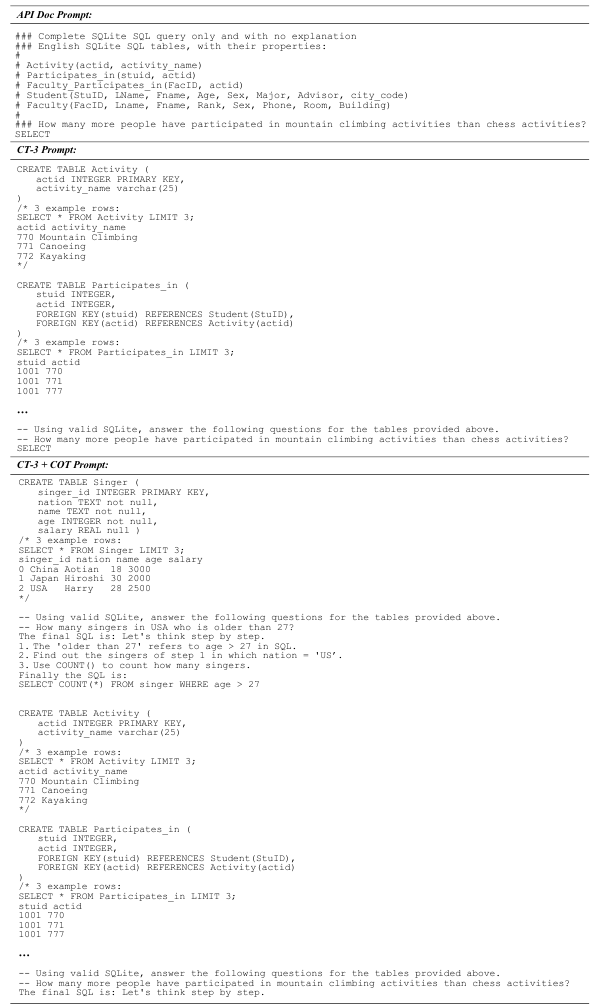}
    \caption{The example of API Doc prompt, CT-3 prompt, and CT-3+COT prompt.}
    \label{fig:prompt-api-doc}
\end{figure}

% \twocolumn

\clearpage

\section{Execution Accuracy Algorithm}
\label{App: EX-A}

\begin{algorithm}[H]
\label{algorithm}
\SetAlgoLined
\KwResult{Check if the execution results of p\_sql and g\_sql against db are equivalent}
\KwIn{p\_sql, g\_sql, db}
\BlankLine
\eIf{p\_sql is not valid for execution against db}{
   \Return False\;
   }{
   Connect to the database db\;
   Execute p\_sql and store the results in pred\_res\;
   Execute g\_sql and store the results in gold\_res\;
   Close the database connection\;
   \BlankLine
   \uIf{pred\_res is exactly equal to gold\_res}{
      \Return True\;
      }
   \uElseIf{the number of rows or columns in pred\_res and gold\_res are different}{
      \Return False\;
      }
   \uElseIf{g\_sql contains an outermost ORDER BY clause}{
      Compare sets of columns in pred\_res and gold\_res\;
      \Return True if equivalent, False otherwise\;
      }
   \Else{
      Calculate element frequency of each row and column in both pred\_res and gold\_res\;
      Check if every frequency in pred\_res is present in gold\_res for both rows and columns\;
      \Return True if all frequencies match, False otherwise\;
      }
   }
\caption{Execution Match Check}
\end{algorithm}

% \begin{algorithm*}[H]
% \footnotesize
% \SetKwInput{KwInput}{Input}                % Set the Input
% \SetKwInput{KwOutput}{Output} 
% \DontPrintSemicolon
% \KwInput{prediction SQL execution result $p\_res$, gold SQL execution result $g\_res$, gold SQL $g\_sql$}
% \KwOutput{execution\_match $m$}

% \If{$p\_res$ == $g\_res$}{
%     return True   \tcp*{quick check}
% }
% \BlankLine
% \If{len($p\_res$)!=len($g\_res$) or len($p\_res$[0])!=len($g\_res$[0])}{
%         return False \tcp*{quickly check the size equivalent}
% }
% \BlankLine

% \If{ there is no \texttt{ORDER BY} in the nest level 0 of $g\_sql$}{ 
%         $p\_res$=sorted($p\_res$) \tcp*{sort the row order} 
%         $g\_res$=sorted($g\_res$) \tcp*{sort the row order} 
% }

% $pred\_columns\_list$ = A list stored the columns of $p\_res$

% $gold\_columns\_list$ = A list stored the columns of $g\_res$
% \BlankLine
% \If {Counter($pred\_columns\_list$) != Counter($gold\_columns\_list$)}{
%         return False             \tcp*{compare columns} 
% }
% \Else{
%         return True
% }
% \caption{Execution Match Algorithm}
% \label{alg: EM-Al}
% \end{algorithm*}

\section{Performance w.r.t Different Reasoning}
\label{App:rea-results}
\begin{table}[H]
\centering
\resizebox{\textwidth}{!}{%
\begin{tabular}{lcccccccccccc}
\toprule
\multirow{3}{*}{\begin{tabular}[c]{@{}l@{}}\textbf{Reasoning}\\ \textbf{Types}\end{tabular}} & \multicolumn{6}{c}{\textbf{EN}}                        & \multicolumn{6}{c}{\textbf{ZH}}                        \\ \cmidrule(lr){2-7} \cmidrule(lr){8-13}
 &
  \multicolumn{2}{c}{\textbf{GPT-3.5 + API Doc}} &
  \multicolumn{2}{c}{\textbf{GPT-3.5 + CT-3}} &
  \multicolumn{2}{c}{\textbf{GPT-3.5 + CT-3 + COT}} &
  \multicolumn{2}{c}{\textbf{GPT-3.5 + API Doc}} &
  \multicolumn{2}{c}{\textbf{GPT-3.5 + CT-3}} &
  \multicolumn{2}{c}{\textbf{GPT-3.5 + CT-3 + COT}} \\
  \cmidrule(lr){2-3} \cmidrule(lr){4-5} \cmidrule(lr){6-7} \cmidrule(lr){8-9}  \cmidrule(lr){10-11} \cmidrule(lr){12-13}
  & \textbf{VA}    & \textbf{EX}    & \textbf{VA}    & \textbf{EX}    & \textbf{VA}    & \textbf{EX}    & \textbf{VA}    & \textbf{EX}    & \textbf{VA}    & \textbf{EX}    & \textbf{VA}    & \textbf{EX}    \\ \hline
A                                                                          & 78.78 & 23.02 & 83.81 & 24.46 & 79.50 & 25.54 & 84.17 & 19.06 & 88.85 & 21.58 & 77.34 & 19.42 \\
A+C                                                                        & 86.60 & 14.05 & 85.95 & 13.73 & 78.43 & 15.69 & 89.87 & 12.09 & 91.18 & 16.67 & 76.80 & 16.99 \\
A+H                                                                        & 78.95 & 9.21  & 83.77 & 7.46  & 67.54 & 4.82  & 83.77 & 6.58  & 92.54 & 6.14  & 64.47 & 4.82  \\
A+C+H                                                                      & 85.65 & 4.35  & 82.61 & 5.22  & 73.04 & 3.48  & 86.09 & 2.61  & 92.61 & 3.91  & 70.43 & 4.35  \\ \bottomrule
\end{tabular}%
}
\caption{Performance with respect to different reasoning types.\vspace{-.4cm}}
\label{tab:performance wrt reasoning-full}
\end{table}

\begin{table}[H]
\centering
\resizebox{\textwidth}{!}{%
\begin{tabular}{lcccccccccccc}
\toprule
\multirow{3}{*}{\begin{tabular}[c]{@{}l@{}}\textbf{Reasoning}\\ \textbf{Types}\end{tabular}} & \multicolumn{6}{c}{\textbf{EN}}                        & \multicolumn{6}{c}{\textbf{ZH}}                        \\ \cmidrule(lr){2-7} \cmidrule(lr){8-13}
 &
  \multicolumn{2}{c}{\textbf{GPT-3.5 + API Doc}} &
  \multicolumn{2}{c}{\textbf{GPT-3.5 + CT-3}} &
  \multicolumn{2}{c}{\textbf{GPT-3.5 + CT-3 + COT}} &
  \multicolumn{2}{c}{\textbf{GPT-3.5 + API Doc}} &
  \multicolumn{2}{c}{\textbf{GPT-3.5 + CT-3}} &
  \multicolumn{2}{c}{\textbf{GPT-3.5 + CT-3 + COT}} \\
  \cmidrule(lr){2-3} \cmidrule(lr){4-5} \cmidrule(lr){6-7} \cmidrule(lr){8-9}  \cmidrule(lr){10-11} \cmidrule(lr){12-13}
   & \textbf{VA}    & \textbf{EX}    & \textbf{VA}    & \textbf{EX}    & \textbf{VA}    & \textbf{EX}    & \textbf{VA}    & \textbf{EX}    & \textbf{VA}    & \textbf{EX}    & \textbf{VA}    & \textbf{EX}    \\ \hline
Addition                                                                       & 80.95  & 33.33 & 88.10 &  23.81 & 73.81 & 16.67 & 92.86 & 30.95 & 95.24 & 26.19 & 78.57 & 19.05 \\
Subtraction                                                                    & 72.41 & 15.52 & 80.17 & 22.41  & 75.00 & 23.28  & 81.03 & 16.38  & 87.93  & 13.79 & 80.17 & 13.79 \\
Multiplication                                                              & 84.52 & 26.19 &  85.12 & 26.19  & 81.55  & 28.57 & 85.71 & 20.24  & 89.88 & 25.60 & 74.40  & 20.24 \\
Division                                                                 
& 81.25 & 16.07 & 79.46 & 18.75  & 76.79 & 28.57 & 80.36 & 12.50  & 85.71  & 18.75 & 68.75 & 16.96 \\ \bottomrule
\end{tabular}%
}
\caption{Performance with respect to different arithmetic operations on data with arithmetic reasoning only.\vspace{-.4cm}}
\label{tab:performance wrt ari-full}
\end{table}

\begin{table}[H]
\centering
\resizebox{\textwidth}{!}{%
\begin{tabular}{lcccccccccccc}
\toprule
\multirow{3}{*}{\begin{tabular}[c]{@{}l@{}}\textbf{Reasoning}\\ \textbf{Types}\end{tabular}} & \multicolumn{6}{c}{\textbf{EN}}                        & \multicolumn{6}{c}{\textbf{ZH}}                        \\ \cmidrule(lr){2-7} \cmidrule(lr){8-13}
 &
  \multicolumn{2}{c}{\textbf{GPT-3.5 + API Doc}} &
  \multicolumn{2}{c}{\textbf{GPT-3.5 + CT-3}} &
  \multicolumn{2}{c}{\textbf{GPT-3.5 + CT-3 + COT}} &
  \multicolumn{2}{c}{\textbf{GPT-3.5 + API Doc}} &
  \multicolumn{2}{c}{\textbf{GPT-3.5 + CT-3}} &
  \multicolumn{2}{c}{\textbf{GPT-3.5 + CT-3 + COT}} \\
  \cmidrule(lr){2-3} \cmidrule(lr){4-5} \cmidrule(lr){6-7} \cmidrule(lr){8-9}  \cmidrule(lr){10-11} \cmidrule(lr){12-13}
   & \textbf{VA}    & \textbf{EX}    & \textbf{VA}    & \textbf{EX}    & \textbf{VA}    & \textbf{EX}    & \textbf{VA}    & \textbf{EX}    & \textbf{VA}    & \textbf{EX}    & \textbf{VA}    & \textbf{EX}    \\ \hline
w/o knowledge                                                    
&  86.19 & 9.89 & 84.51 & 10.07  & 76.12 & 10.45  & 88.25 & 8.02 & 91.79 & 11.19 & 74.07 & 11.57 \\
w/ knowledge                                                     
& 83.58  & 9.89 & 87.13 & 12.31  & 74.81 & 13.43 & 85.07 & 9.70 & 87.69 & 12.13 & 73.32 & 13.62 \\
\bottomrule
\end{tabular}%
}
\caption{Performance for questions needed commonsense reasoning with and without explicit knowledge input.\vspace{-.4cm}}
\label{tab:performance wrt commonsense-full}
\end{table}

\begin{table}[H]
\centering
\resizebox{\textwidth}{!}{%
\begin{tabular}{lcccccccccccc}
\toprule
\multirow{3}{*}{\begin{tabular}[c]{@{}l@{}}\textbf{Reasoning}\\ \textbf{Types}\end{tabular}} & \multicolumn{6}{c}{\textbf{EN}}                        & \multicolumn{6}{c}{\textbf{ZH}}                        \\ \cmidrule(lr){2-7} \cmidrule(lr){8-13}
 &
  \multicolumn{2}{c}{\textbf{GPT-3.5 + API Doc}} &
  \multicolumn{2}{c}{\textbf{GPT-3.5 + CT-3}} &
  \multicolumn{2}{c}{\textbf{GPT-3.5 + CT-3 + COT}} &
  \multicolumn{2}{c}{\textbf{GPT-3.5 + API Doc}} &
  \multicolumn{2}{c}{\textbf{GPT-3.5 + CT-3}} &
  \multicolumn{2}{c}{\textbf{GPT-3.5 + CT-3 + COT}} \\
  \cmidrule(lr){2-3} \cmidrule(lr){4-5} \cmidrule(lr){6-7} \cmidrule(lr){8-9}  \cmidrule(lr){10-11} \cmidrule(lr){12-13}
   & \textbf{VA}    & \textbf{EX}    & \textbf{VA}    & \textbf{EX}    & \textbf{VA}    & \textbf{EX}    & \textbf{VA}    & \textbf{EX}    & \textbf{VA}    & \textbf{EX}    & \textbf{VA}    & \textbf{EX}    \\ \hline
Hypothetical                                                   
&  82.31 & 6.77 & 83.19 & 6.33  & 70.31 & 4.15 & 84.93 & 4.59 & 92.58 & 5.02 & 67.47 & 4.59  \\
Factual                                                    
&  82.53 & 17.25 & 83.84 & 18.12  & 79.48 & 20.09 & 87.77 & 15.5 &  89.52 & 17.47  & 79.48 & 16.81 \\
\bottomrule
\end{tabular}%
}
\caption{Performance comparison for hypothetical questions and corresponding factual questions.\vspace{-.4cm}}
\label{tab:performance wrt hyp-full}
\end{table}

\newpage
\section{Archer Examples}
\label{app:archer-ex}

\begin{figure}[htbp]
    \centering
    \includegraphics[width=\textwidth]{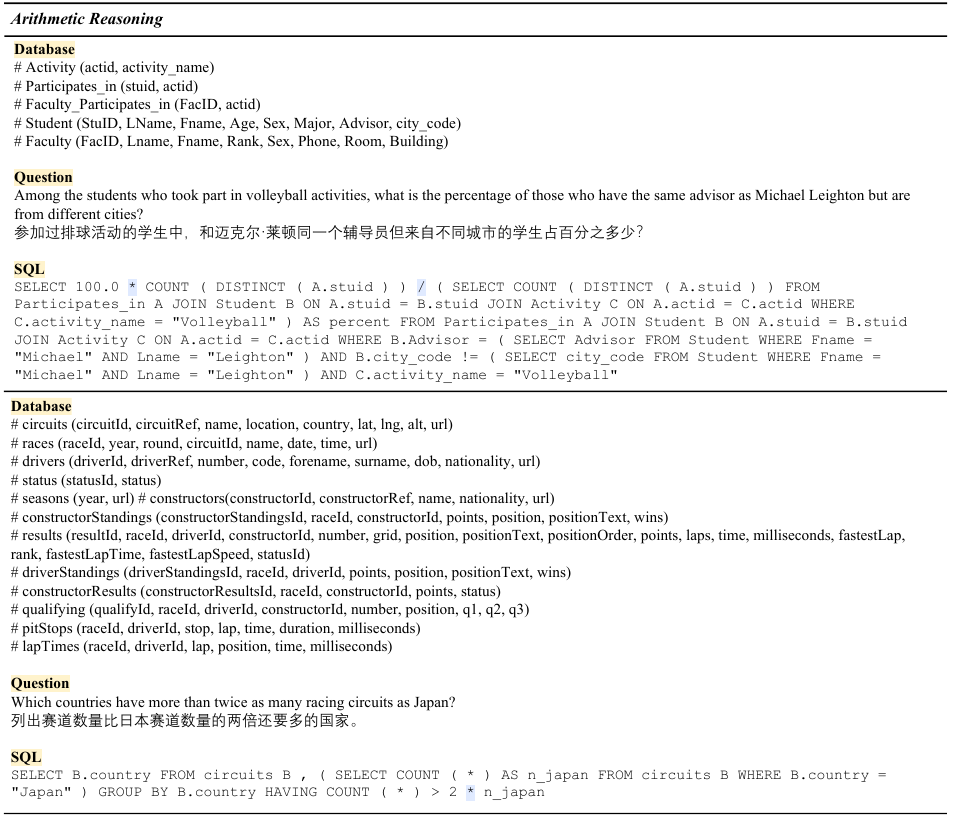}
    \caption{The example of Archer data requiring Arithmetic Reasoning.}
    \label{fig:ari_ex}
\end{figure}

\begin{figure}[htbp]
    \centering
    \includegraphics[width=\textwidth]{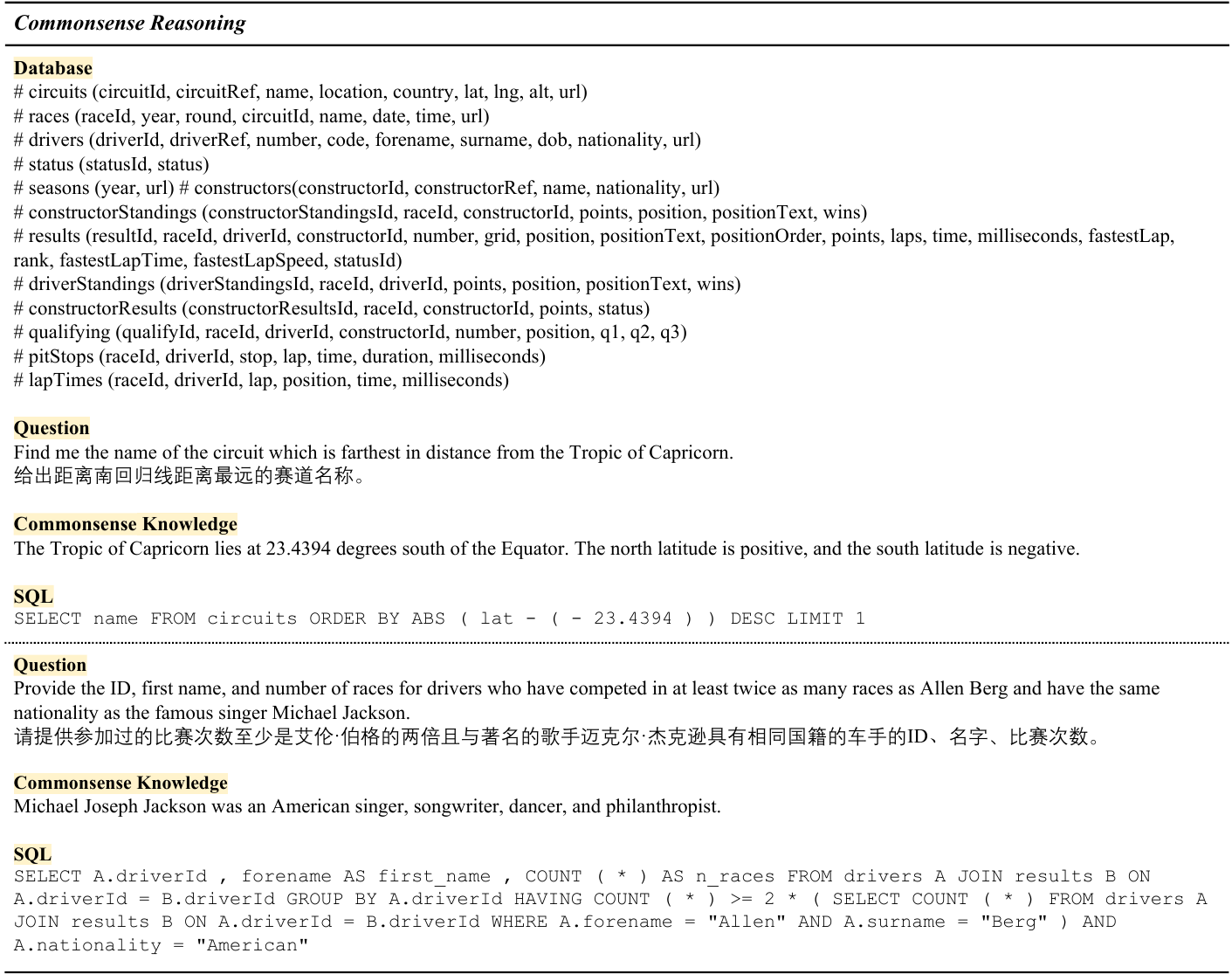}
    \caption{The example of Archer data requiring Commonsense Reasoning.}
    \label{fig:com_ex}
\end{figure}

\begin{figure}[htbp]
    \centering
    \includegraphics[width=\textwidth]{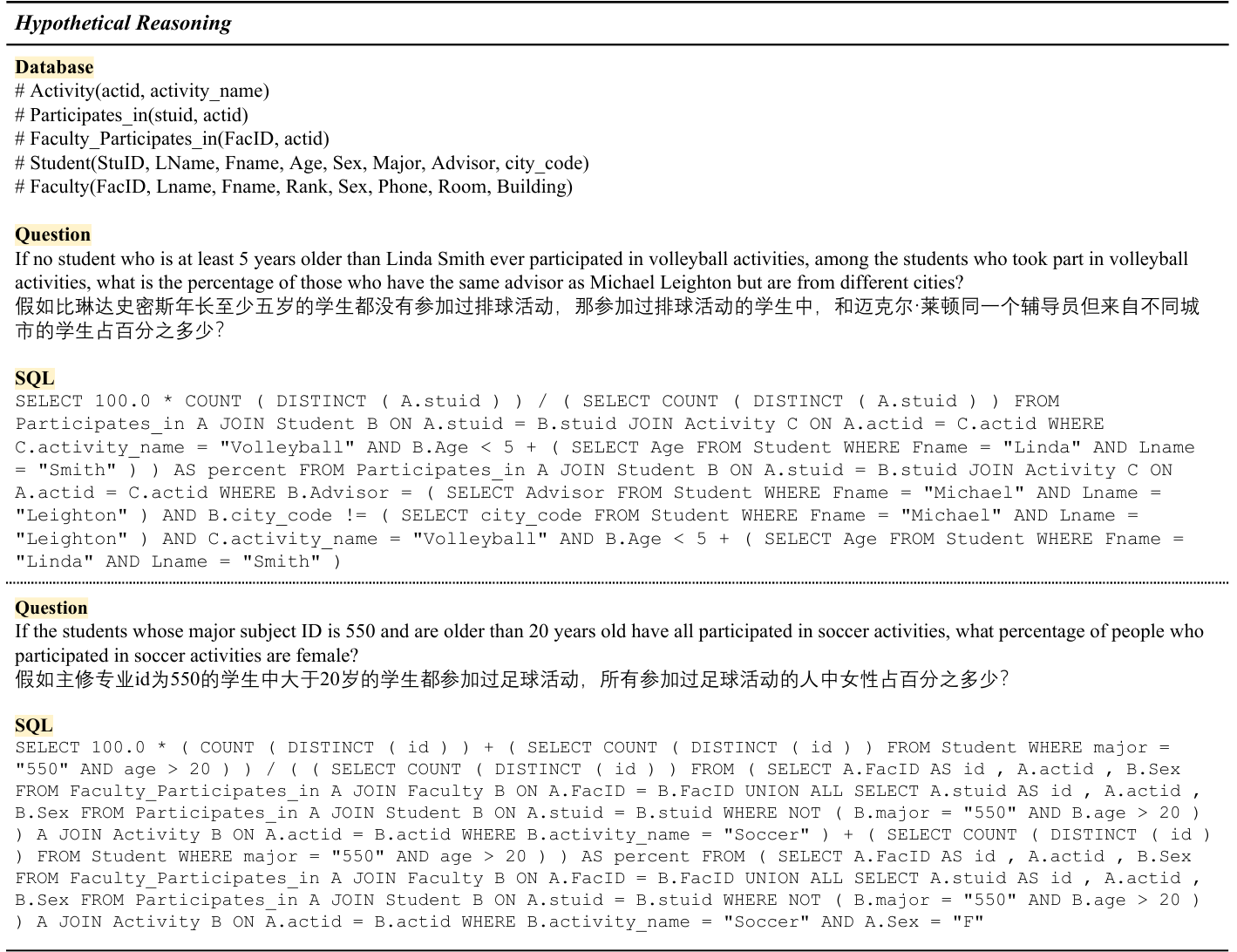}
    \caption{The example of Archer data requiring Hypothetic Reasoning.}
    \label{fig:hyp_ex}
\end{figure}

\end{document}